\documentclass[10pt,letterpaper]{article}
\usepackage[top=0.85in,left=2.75in,footskip=0.75in]{geometry}

\usepackage{amsmath,amssymb}

\usepackage{changepage}

\usepackage[utf8x]{inputenc}

\usepackage{textcomp,marvosym}

\usepackage{cite}

\usepackage{nameref}
\usepackage{hyperref}
\usepackage{float}
\usepackage[right]{lineno}
\usepackage{rotating}
\usepackage{microtype}
\DisableLigatures[f]{encoding = *, family = * }

\usepackage[table]{xcolor}

\usepackage{array}

\newcolumntype{+}{!{\vrule width 2pt}}

\newlength\savedwidth

\raggedright
\usepackage[aboveskip=1pt,labelfont=bf,labelsep=period,justification=raggedright,singlelinecheck=off]{caption}
\renewcommand{\figurename}{Fig}

\makeatletter
\renewcommand{\@biblabel}[1]{\quad#1.}
\makeatother

\usepackage{lastpage,fancyhdr,graphicx}
\usepackage{epstopdf}
\fancyheadoffset[L]{2.25in}
\fancyfootoffset[L]{2.25in}
\usepackage{xcolor}

\begin{document}
\vspace*{0.2in}

\begin{flushleft}
{\Large
\textbf\newline{Assessing the Spatial Structure of the Association between  Attendance at Preschool and Children's Developmental Vulnerabilities in Queensland, Australia} 
}
\newline
\\
Wala Draidi Areed\textsuperscript{1*},
Aiden Price\textsuperscript{1},
Kathryn Arnett\textsuperscript{2},
Helen Thompson\textsuperscript{1},
Reid Malseed\textsuperscript{2}
Kerrie Mengersen\textsuperscript{1},
\\
\bigskip
\textbf{1}School of Mathematical Science, Center for Data Science, Queensland University of Technology, Queensland, Australia.
\\
\textbf{2}Children's Health Queensland, Queensland, Australia.
\\
\bigskip

%
%





* w.areed@qut.edu.au

\end{flushleft}
\section*{Abstract}
Demographic and educational factors are essential, influential factors of early childhood development. This study aimed to investigate spatial patterns in the association between attendance at preschool and children's developmental vulnerabilities in one or more domain(s) in their first year of full-time school at a small area level in Queensland, Australia. This was achieved by applying geographically weighted regression (GWR) followed by $K$-means clustering of the regression coefficients. Three distinct geographical clusters were found in Queensland using the GWR coefficients. The first cluster covered more than half of the state of Queensland, including the  Greater Brisbane region, and displays a strong negative association between developmental vulnerabilities and attendance at preschool. That is, areas with high proportions of preschool attendance tended to have lower proportions of children with at least one developmental vulnerability in the first year of full-time school. Clusters two and three were characterized by stronger negative associations between developmental vulnerabilities, English as the mother language, and geographic remoteness, respectively. This research provides
evidence of the need for collaboration between
health and education sectors  in specific regions of Queensland to update
current service provision policies and to ensure
holistic and appropriate care is available to support children with developmental vulnerabilities.

\section*{Introduction}
\begin{figure}[h!]
    \centering
   \includegraphics[scale=0.5]{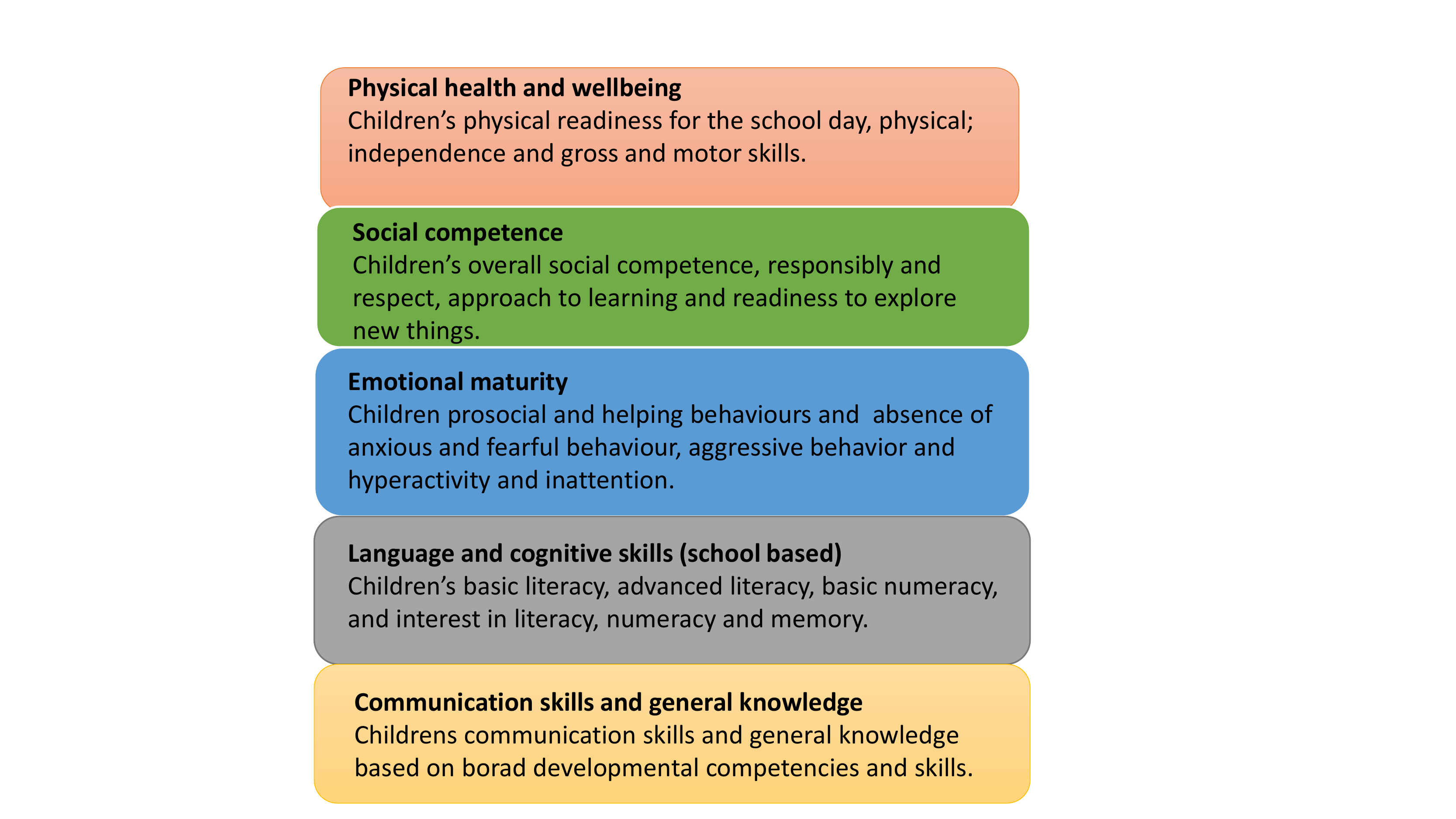}
    \caption{Early childhood development domains defined by AEDC.}
    \label{fig:my_label1}
\end{figure}

The first five years of a child’s life, commonly referred to as early childhood \cite{hoff2009language} have a significant long-term impact on later development, even into adulthood \cite{berlin1998effectiveness}. As a result, early childhood health and development assessments are of great interest to communities and government agencies to facilitate targeted early intervention strategies which can allow children to reach their maximum developmental potential \cite{to2001biological}. Many countries, including Australia, are increasingly using national progress indicators of early childhood development to aid in these assessments.\\
In Australia, these progress indicators are collected as part of the population-based Australian  Early Development Census (AEDC), conducted every three years since 2009 \cite{janus2007early}. Using the 2009 census results as a benchmark, scores ranging between 0 and 10 are calculated  for each child for each of five development domains and children are classified as developmentally vulnerable (less than 10th percentile), at-risk (between the 10th
and 25th percentile), or on track (above the 25th percentile) in each domain.\\ The five developmental domains are physical health and well-being, social competence, emotional maturity, language and cognitive skills (school based), and communication skills and general knowledge; see Fig ~\ref{fig:my_label1} \cite{shonkoff2009investment}.  The scores on these developmental domains are publicly available on the AEDC data explorer at the small area level; see section \nameref{data} for details.  
\\
While the focus of the AEDC is on the specific milestones in childhood development within each of the domains \cite{council2008national}, there is potential to gain additional insight into each of the early childhood development domain by assessing the spatial relationship of these data with socio-demographic factors and educational factors. An important educational factor is attendance at preschool, defined as structured, play-based education provided to children prior to school entry by a qualified early childhood teacher \cite{council2008national}. Preschool provides young children with rich learning environments that can enhance their cognitive, physical, social, and emotional development \cite{hall2013can}. Attending preschool may also improve the chance of effective school transitions, with long-term implications for future academic and occupational success. As a result, policymakers are becoming more interested in the potential of preschool to improve developmental readiness for school. In line with this viewpoint, recent national reform programs in Australia have aimed to encourage preschool attendance by providing universal access to a preschool program in the year before the start of school \cite{council2008national}. Despite these efforts, however, not all children attend preschool. In 2021, 85\% of all 4-year-old and 22\% of all 5-year-old children were enrolled in preschool programs in Australia.  \cite{gov}. \\
Understanding geographic variation in preschool and developmental vulnerability in the first year of full-time school is important for communities, health managers and policymakers.
This paper uses data available at a small area level (statistical area level 2, SA2)  to investigate spatial patterns and clusters for the proportion of vulnerable children within the AEDC domains \cite{collier2020inequalities}. The analysis concentrates on the state of Queensland, Australia, and on the association between developmental vulnerabilities and attendance at preschool, taking into account socio-demographic factors (country of birth, English is the primary language, remoteness, The Index of Relative Socio-economic Disadvantage); see section \nameref{data}. In addition, The investigation of whether geographical and educational factors have similar effects across a study region has been conducted. In this study, the results for vulnerability on one or more domain(s) as a summary of the developmentally vulnerable provided by AEDC have shown; however, the analyses are carried out for all the AEDC developmental domains and are reported in supplementary material \nameref{S3}. 
\\  Various statistical approaches can be used to discover clusters in the aggregated data \cite{buchin2012processing, rogerson2008statistical,waller2004applied}. These approaches include the spatial scan statistic \cite{kulldorff1997spatial,kulldorff1995spatial}, the Geographical Analysis Machine \cite{charlton2006mark}, Bayesian varying coefficients models for areal data \cite{gamerman2003space}, and penalized local polynomial models  \cite{wang2019penalized}. In addition, some studies have employed varying coefficient regression models based on spatial cluster frameworks. For example,  Lawson \cite{lawson2014prior} proposed an approach
that provides the grouping of regression coefficients directly when the number of groups is known a priori. Lee \cite{lee2017cluster}
proposed a spatial cluster detection method for regression coefficients, which directly identifies an unknown number
of spatial clusters in the regression coefficients via hypothesis testing and the construction of spatially varying coefficient regression based on detected spatial clusters. More
recently, Lagona \cite{lagona2020model} proposed to estimate space-varying effects on the regression coefficients by exploiting a multivariate
hidden Markov field and using an expectation-maximization algorithm and composite likelihood methods.\\
A method that considers non-stationary variables and models the local relationships between these predictors and an outcome of interest is the geographically weighted regression algorithm (GWR) suggested by Brunsdon, Fotheringham, and Charlton \cite{fotheringham1998geographically}. This is a local spatial technique that addresses both spatial heterogeneity  and spatial dependence (i.e. spatial dependence
and spatial autocorrelation), which generates locally weighted regression coefficients that vary geographically \cite{fotheringham1998geographically,brunsdon1998geographically,leong2017modification}. 
The output of the GWR is a set of regression coefficients for each location. 

In order to identify spatial patterns in the data, a combination of the GWR coefficients with a  $K$-means cluster algorithm \cite{yadav2013review} employed for the local regression coefficients to identify groups of SA2 areas that are similar with respect to the relationships between attendance at preschool and developmental vulnerabilities in the first year of full-time school. \\GWR has been used in a number of real-world applications that show spatial heterogeneity in estimated covariate effects \cite{acharya2018modeling,black2014ecological,ribeiro2018modelling,sultana2018household,wang2017analyzing,wu2017prediction,yang2012understanding}. This combination of GWR and $K$-means has been deployed in many fields including ecology \cite{windle2010exploring,ye2020investigating}, economics \cite{kopczewska2021spatio,hernandez2021using}, health \cite{zhang2021space, wang2020spatial}, environment \cite{cullen2021use, li2020spatially,tutmez2012evaluating,zhao2020compactness}, social media \cite{nicholson2019spatial}, education \cite{sacco2021spatial},
and transportation \cite{bao2018spatial}.
To the best of our knowledge, no studies have
been published that consider this type of spatial clustering for children's AEDC domains. Although a previous study investigated the association between early life risk factors and children's developmental vulnerabilities at age 5 using latent class analysis, spatial factors were not taken into account \cite{taylor2020associations}. The results presented in this paper provide an opportunity to shed more light on the development of children in Queensland.
\section*{Materials and methods}
\subsection*{Case Study Area} \label{area}
\begin{figure}
     \centering
    \includegraphics[height=18cm, width=15cm]{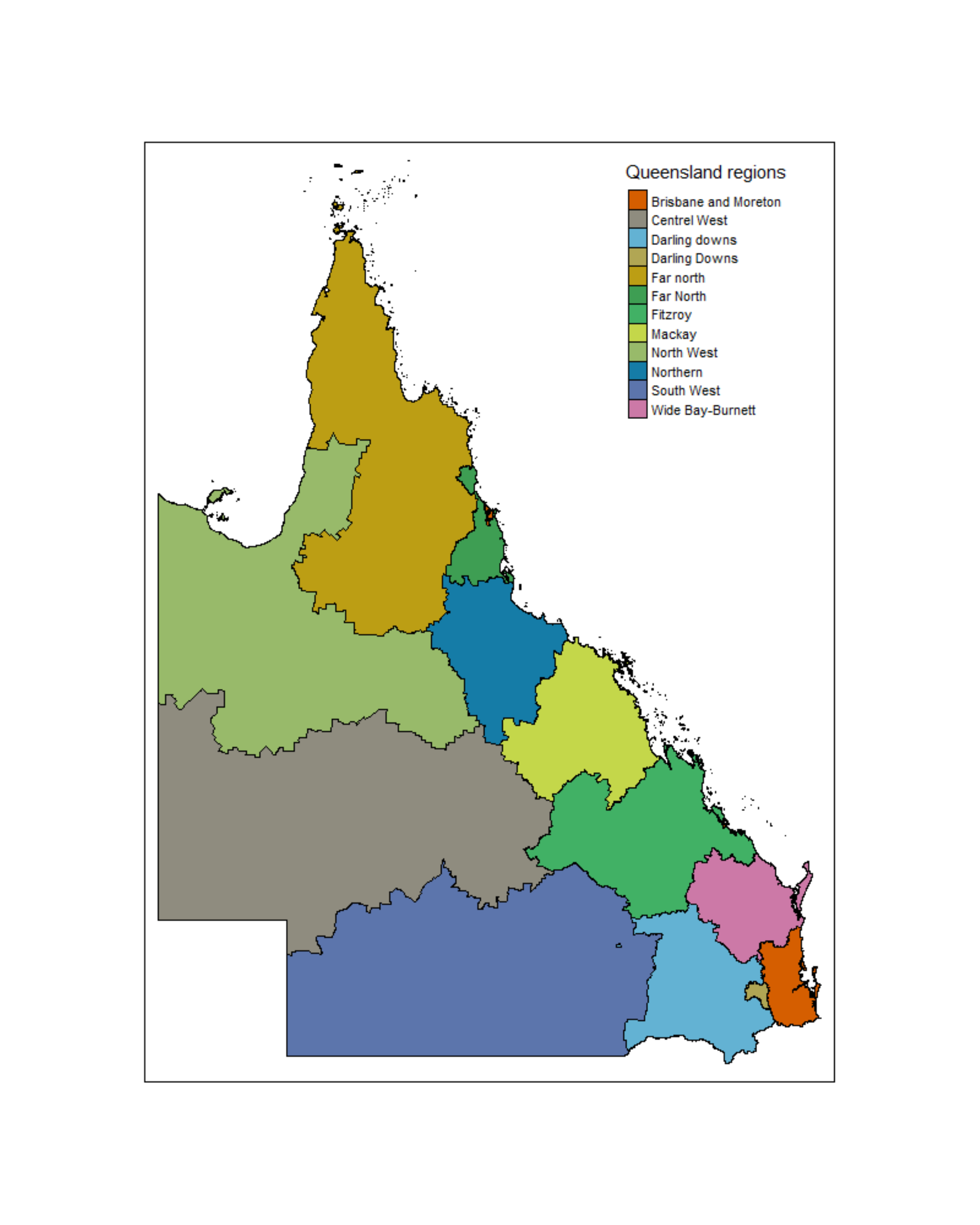}
     \caption{The structure of the main regions in Queensland, some notable regions include:
     1) Southeast Queensland (SEQ) is home to more than 70\% of the state's population. It contains two statistical regions, Greater Brisbane and Moreton.
2) Darling Downs in the state's inland south-east, which   includes the city of Toowoomba
3) South West Queensland in the state's inland south-west.
4) Central West in the state's inland central-west.
5) Wide Bay-Burnett is located north-east of the Darling Downs and north of the Sunshine Coast
6)Central Queensland, which includes Fitzroy and Mackay
7) North Queensland on the state's northern coastline, which   includes the city of Townsville.
8) North West in the state's inland north-west of Queensland and includes the city of Mount Isa.
9) Far North in the state's extreme northern coastline and also includes the city of Cairns.\\
}
     \label{fig:my_label}
 \end{figure}

The state of Queensland, Australia, is divided geographically into nine large regions (\figurename{\ref{fig:my_label}}) and 528 non-overlapping statistical area level 2 (SA2) regions (according to the Australian Statistical Geography Standard  (ASGS) 2016 boundaries of the Australian Bureau of Statistics(ABS)). SA2 regions are medium-sized general-purpose regions that reflect a socially and geographically region integrated community \cite{ABS}. SA2 is the smallest geographic area at which ABS non-census and intercensal data are publicly released.\\
 Brisbane is Queensland’s state capital city with a population of over 2.58 million \cite{region}, which ranks as the 3rd most populated city in Australia. The Greater Brisbane region is located in south-eastern Queensland and includes 236 SA2 areas.
 \subsection*{Case Study Data} \label{data}
As described previously, the development domains data for this study were obtained from the Australian Early Development Census. The AEDC collects comprehensive statistics on children in Australia every three years. Teachers conduct a census of their students in their first year of full-time school. The data are used to establish scores for each of five domains: physical health
and well-being (Physical), social competence (Social), emotional maturity (Emotional), language and cognitive skills (Language), and
communication skills and general knowledge (Communication), \figurename{ \ref{fig:my_label1}}. Each child is given a score between zero and ten for each domain, based on the cut-offs established as a baseline in 2009. When monthly age differences are considered, children who fall below the 10th percentile in a domain are labelled ``developmentally vulnerable". AEDC also derives two additional domain indicators: vulnerable on one or more domains (Vuln 1) and vulnerable on two or more domains (Vuln 2). \\
A range of factors are associated with the child's early development. These can be broadly classed as factors related to the child, the mother, the family and the built environment. Factors related to the child include: Indigenous status, low birth weight, number of siblings, and country of birth \cite{brinkman2012jurisdictional}. Maternal risk variables include teenage mother at birth of the child (less
than 20 years), smoking in pregnancy, and alcohol use in pregnancy \cite{christian2015influence}. Other family risk factors include: non-English speaking parents, single parents, moved house in last 12 months, main carer and parent education \cite{guhn2016associations}. Finally, built environment factors include: home yard area, distance to the nearest park, distance to nearest family support service, distance to nearest playgroup venue, distance to nearest kindergarten, residential density \cite{christian2015influence,webb2017neighbourhood,walker2004national}.\\
 Covariates considered for inclusion in this study were selected based on the existing literature and AEDC website \cite{AEDC}
regarding their role as potential confounders or important contextual
variables at area level, in the associations between geographic and educational factors on children's development  \cite{goldfeld2016role,guthridge2016early,moore2015early}. The final list of covariates used in this study comprised attendance at preschool (Preschool), English as the mother language (English), Australia as the country of birth (Australia), the Socio-Economic Indexes For Areas (IRSD), and remoteness ( major city, inner-regional, outer-regional, remote, and very remote). There are 294 SA2 regions classified as major cities,  113 SA2 areas classified as inner regional,  96 SA2 areas classified as outer regional, 11 SA2 areas classified as remote and 14 SA2 areas classified as very remote.\\
The IRSD index (The Index of Relative Socio-economic Disadvantage) is scored on a scale of one to five. A low score indicates that the region as a whole is at a disadvantage; this includes many low-income families, many people without qualifications, or low-skill occupations.\\

This study used the latest publicly available data from the 2018-2019 census from AEDC. The participation rate in Queensland was with 98.1\% of eligible children
represented in the dataset in 2018. Between 3\% and 6\% of the data were missing variables in the dataset. Spatial neighbourhood averages were used to impute missing continuous data. The highest frequency neighbourhood category was used for categorical data. Due to the lack of contiguous neighbours in two cases, missing values for two islands could not be filled. As a result, the scope of this study's investigation was reduced to the remaining 526 SA2 regions.
\subsection*{Spatial autocorrelation analysis}
Moran's I was used to investigate global spatial autocorrelation for each type of vulnerability domain \cite{cliff1981spatial}. This paper defines neighbours as spatial units (SA2s) that share an edge or vertex. This classification of neighbours is known as the queen criterion \cite{cliff1981spatial}. Moran's I spatial autocorrelation algorithm takes values between -1 and 1. Coefficients between 0 and 1 show positive spatial autocorrelation; negative coefficients between 0 and -1 imply different neighbouring values, and coefficients approaching 0 indicate weak or no spatial autocorrelation \cite{paradis2009moran}. For more details, see \nameref{support}.

\subsection*{Geographically Weighted Regression}
The geographically weighted regression (GWR) model \cite{brunsdon1996geographically}, which is an extension of ordinary least squares (OLS), was adopted for this study. A weighted spatial matrix, which depicts local geographic interactions, is produced using spatial kernel functions.  The weight (W) is a matrix of weights specific to location $i$ such that observations nearer to $i$ are given greater weight than observations further away \cite{fotheringham1998geographically}.   \\
Given a response vector $\underline{y}=\{y_1,y_2,...,y_n\}$ and a $n \times p$ matrix of covariates $X$, the GWR model is written as:
\begin{equation}
    Y_i= \beta_0(u_i,v_i)+\sum_{k=1}^{p} \beta_{k}(u_i,v_i) X_{ik} +\epsilon_i \quad \quad i=1,2,3,...,n,
\end{equation}
where $Y_i$ is the dependant attributes at each location $i$ and $\underline{y}$= $(y_1,y_2,...,y_n)^\top$, $X_{ik}$ is the $k$-th covariate at location $i$, $(u_i,v_i)$ denotes the coordinates of point $i$ in space (longitude and latitude), $ \beta_0(u_i,v_i),...,\beta_k(u_i,v_i)$ are model parameters, and $\epsilon_i$ is the random error at location $i$ with mean zero and variance $\sigma^2 I$, where $I$ is the identity matrix \cite{fotheringham1998geographically}. GWR thus allows the coefficients to vary spatially, with the estimated values of the coefficient values given by:
\begin{equation}
      \hat{\beta} (i)=[X^{T} W(i) X]^{-1} X^{T} W(i) Y,
\end{equation}
where $W(i)$ denotes the spatial weight matrix for location $i$.  In this paper, different kernel functions between observations and local regressions were used to find $W$. The fixed Gaussian  kernel was adopted, given by:
\begin{equation}
    w_{ij}=exp(-(d^{s}_{ij})^2/ b^2),
\end{equation}
where the $b_i$ represents the bandwidth which is the radius of a
point to the extent to which the point influences that form a circle, $d_{ij}$ is the euclidean distance \cite{fotheringham2015geographical}. An adaptive bi-square  was also considered with the expression \cite{fotheringham1991modifiable}, 
\begin{equation}
 w_{ij}=\begin{cases} 
      [1-(d_{ij}^s/b_i)^2]^2 & \text{if} \quad d_{ij}^s < b_i,\\

      0 & \text{otherwise},
   \end{cases}
\end{equation}
\\Cross-validation was used to search for the optimal value of $b_i$ bandwidth with the optimal solution returning the smallest model residuals based on a given model specification \cite{yu2007modeling}. Finally, a fixed kernel approach with the same bandwidth at all observation locations was used to explore the effects on the results. The performance of the models was evaluated using quasi-global $R^2$, and goodness of fit $R^2$ \cite{liu2008goodness,smart2021quantifying}.
\subsection*{Clustering of GWR coefficients}
The inferential capability of GWR was extended by clustering together locations with similar sets of parameter values. This synthesises the often vast amount of output created by the GWR model
and aids the interpretation of multiple parameter estimate maps.\\
In this study, $K$-means clustering was employed  \cite{huang1998extensions}, and the results were spatially visualised to 
investigate the spatial clusters based on the GWR coefficients. In the 
$K$-means clustering approach, the similarity between a pair of objects was defined by Euclidean distance, and the objects were partitioned into $K$ clusters, such that the within-cluster sum of squares is minimised \cite{wang2020spatial}. In addition, the silhouette score was used to evaluate the clustering results \cite{rousseeuw1987silhouettes}. A silhouette score close to 1 suggests that the objects are close to the centroid of their respective clusters. In contrast, a score close to 0 indicates that the objects are outliers. The algorithm proceeds as follows. 1) Define the number of clusters $K$. 2) Randomly select $K$ data points within each cluster as the cluster centroids. 3) Assign data points to the closest cluster centroid. 4) Recompute the cluster centroids. 5) Repeat steps 3) and 4) until either the centroids do not change, or the maximum number of iterations is reached. Two sets of cluster analyses were performed using the GWR coefficients. The first set of analyses encompassed all 526 SA2 regions in Queensland (excluding two islands as described previously). Each domain was analysed separately using GWR and $K$-means. The results for Vuln 1 are reported below.\\
 Since most of the population of QLD residents lives in the southeast corner, a separate cluster analysis was performed to understand the spatial patterns in the 236 SA2 areas in the Greater Brisbane region (see \nameref{S2}). In addition, a comparison between the five domains with the cluster of the highest  proportion of attendance preschool coefficient has been discussed. 
\subsection*{Computation}
The analyses were carried out using R statistical software version R-4.1.3 packages (spgwr) \cite{bivand2017package} and (tmap) \cite{tennekes2018tmap}  for the GWR, and mclust  \cite{fraley2012package}, and factoextra \cite{kassambara2017package} packages for the $K$-means algorithm. Spatial distributions of the clusters were visualised with maps using tmap \cite{tennekes2018tmap}, and ggplot2 \cite{wickham2016package} packages in R, and the ArcGIS Pro 2.9.1 software \cite{lloyd2010spatial}.
\section*{Results}
\subsection*{Spatial characteristics of children's vulnerability on one or more domain(s)}
\begin{figure}
  \centering
   \includegraphics[scale=0.4]{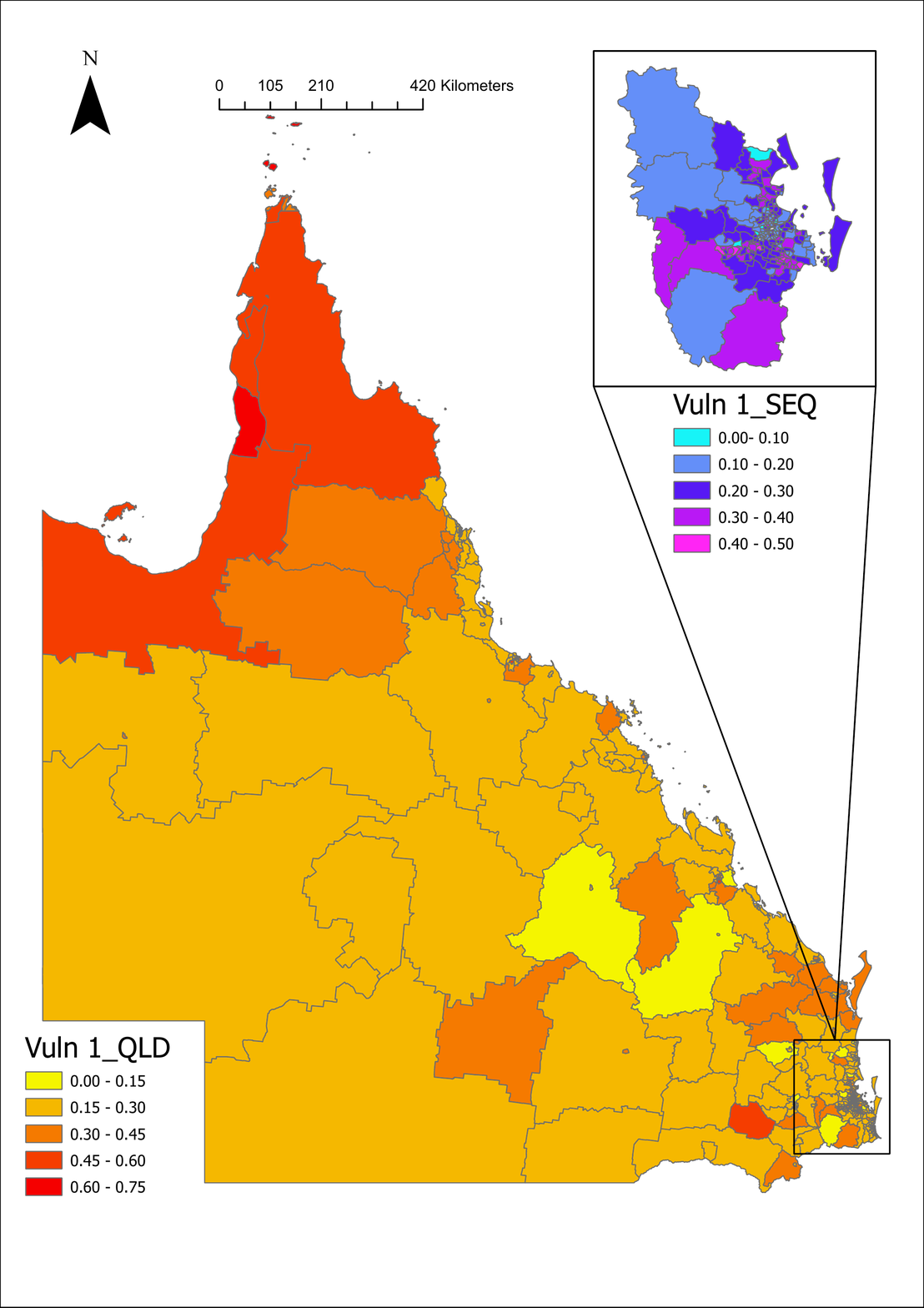}
    \caption{ Spatial distribution of the proportion of developmental vulnerable on one or more domain(s) at the SA2 level in Queensland, 2018-2019.}
     \label{fig:my_label 1}
  \end{figure}

The global Moran’s I for Vuln 1
was 0.36 (p-value =1.0E-04), indicating a highly significant
positive spatial autocorrelation between SA2 areas.
Spatial differences in Vuln 1 were observed (\figurename{ \ref{fig:my_label 1}}),
Vuln 1 is declining from the north to the south. In
general, small values of Vuln 1 (less than 0.30) were located in the southeastern  and central parts of Queensland.
In contrast, high Vuln 1 values (greater than 0.45) were mainly in the northwest and far north of Queensland.
  
  The GWR analysis of Vuln 1 produced a quasi-global GWR $R^2$ of 0.42 and local $R^2$
values  between 0.37 and 0.43 (\nameref{support}).

\begin{figure}[h!]
    \centering
    \includegraphics[scale=0.6]{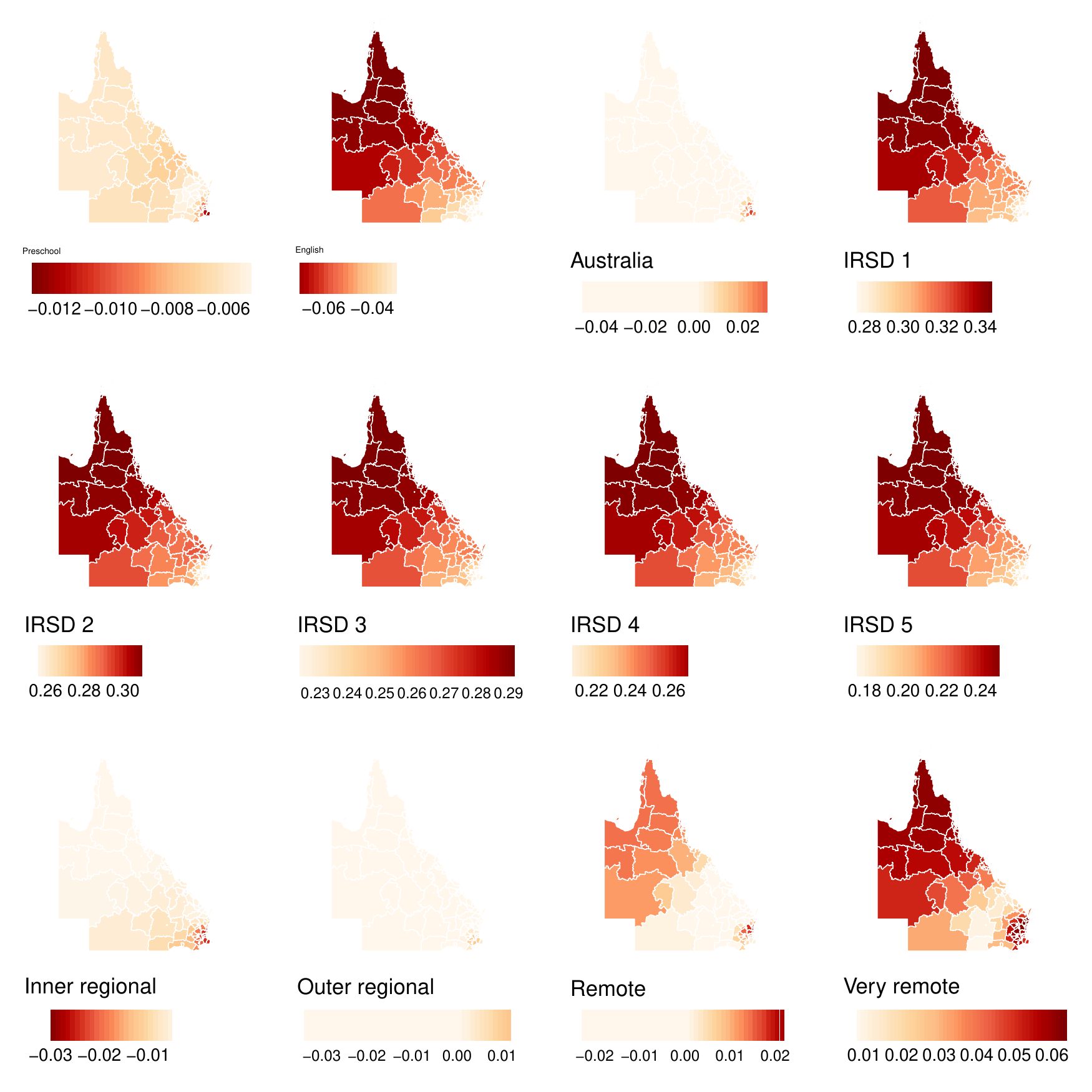}
    \caption{Spatial distribution of GWR coefficients in Queensland.}
    \label{fig:my_labelco}
\end{figure}

 Notably, the SA2 areas with relatively high local $R^2$ were mainly
concentrated in the northwest and far north of Queensland, while low local $R^2$ values were found in the southwest and some parts of the Darling Downs regions. The Moran's I values for the standardized residuals of the GWR models was 0.17 (p-value $<$0.001), indicating substantial spatial variation remaining even after the regression analysis.\\
The GWR  regression coefficients are summarised in Table \ref{tab:my-table1}. The number of SA2 areas with statistically significant GWR coefficients and corresponding maps of these areas are provided in the \nameref{S3}.
\begin{table} 
\centering
\caption{Summary statistics for the GWR model coefficients with global p-values from an ordinary least square regression model (OLS), with major cities as a baseline for remoteness factor. }
\scalebox{0.8}{
\begin{tabular}{cccc}
   \hline
Explanatory variables               & Mean GWR coefficient     & Range GWR coefficient & Global p-value \\ \hline
Preschool & -0.010 & [-0.013, -0.005] &  0.040                           \\
English            & -0.041 &  [ -0.074, -0.028] &        0.007                                 \\
Australia       &0.002 &   [-0.046 0.035] &    0.695                     \\

IRSD (Quintile 1)                           & 0.301 &   [ 0.276, 0.345] &       $<$ 2e-16                                 \\
IRSD (Quintile 2)                            & 0.275&   [0.254, 0.310] &    $<$ 2e-16                                    \\
IRSD (Quintile 3)                            & 0.249&    [0.224, 0.291] &   $<$ 5.00e-15                                  \\
IRSD (Quintile 4)                            & 0.230 &   [0.204, 0.276]  &     5.45e-13                                     \\
IRSD (Quintile 5)                          &0.202 &  [0.175, 0.251]  &      6.86e-11 
\\
Remoteness (Inner regional)                             & -0.019&  [ -0.031, -0.004] &     0.406                                \\
Remoteness (Outer regional)                             &  -0.012 &  [-0.035,  0.011] &     0.428                                 \\
Remoteness (Remote)                              &0.008 &  [-0.024,  0.021] &    0.616                             \\
Remoteness (Very remote)                             &0.049 &  [0.008, 0.063] &   0.0004                                \\

Quasi-global  $R^2 $       &0.42 &   &                                   
                          \\ \hline
\end{tabular}}
\label{tab:my-table1}
\end{table}
Fig \ref{fig:my_labelco} shows the spatial distribution of the GWR coefficients. GWR coefficients corresponding to the proportion of attendance at preschool were largely negative in southeast Queensland,
indicating a substantively negative relationship between the proportion of vulnerability on one or more domain(s) and  the proportion of attendance at preschool at the SA2 level;  the proportion of preschool increases implies the proportion of vulnerable decreases. As noted in section \nameref{area}, this is the most populous region in the state and contains the capital city, Brisbane. In addition, the GWR coefficients were largely negative for the proportion of children with English as the mother language in the northwest and the far  north regions. In contrast, positive GWR coefficients were found for low socioeconomic status (IRSD) in the far north, north-west and central west of Queensland. Finally,
compared to the Baseline inner cities for the remoteness factor, a negative relationship with Vuln 1 for the region classified as inner regions in the south-north of Queensland has been found, and a positive relationship for remote regions in the northwest and far north of Queensland has been noticed. 

\subsection*{Cluster analysis of GWR coefficients}
\begin{figure}[h!]
  \centering
    \includegraphics[width=0.9\textwidth]{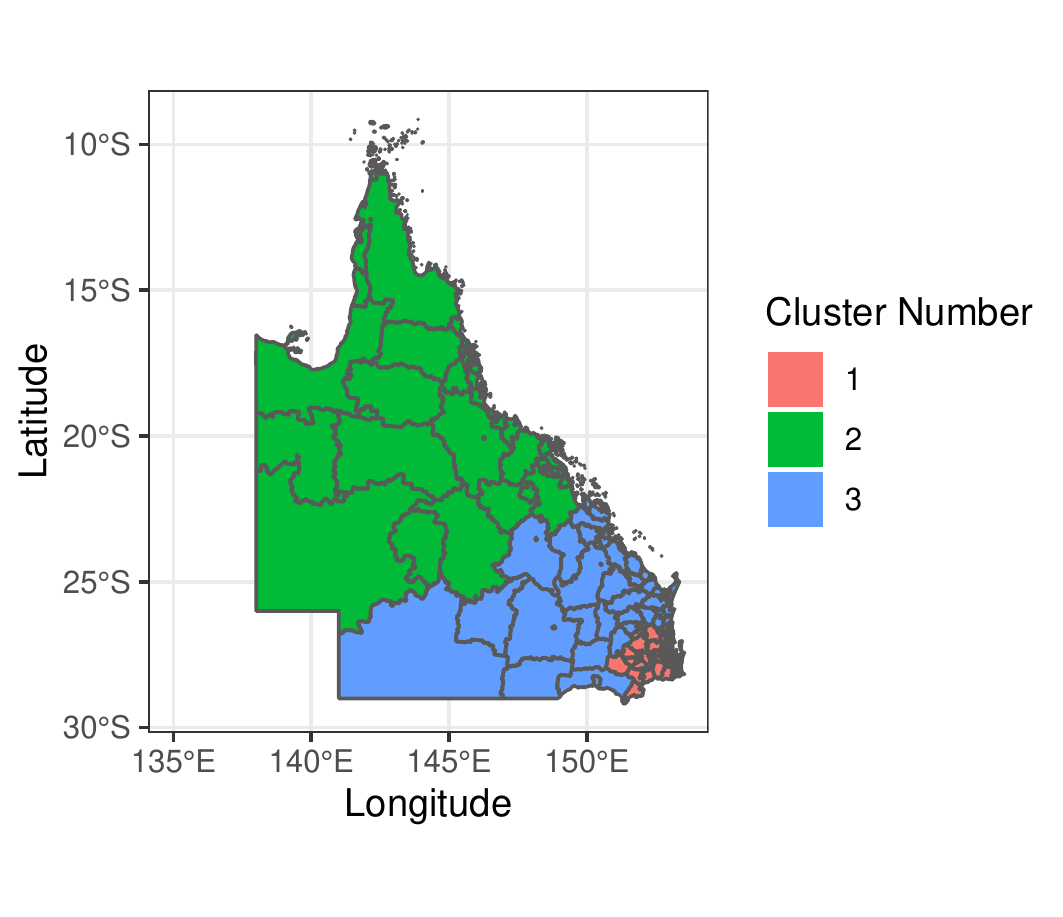}
    \caption{Spatial distribution of clusters of GWR coefficients at SA2 level in Queensland. Insight into the SEQ corner is given in the \nameref{S2}.}
     \label{fig:my_label5}
  \end{figure}

\begin{figure} [h!]
    \centering
    \includegraphics[width=0.9\textwidth]{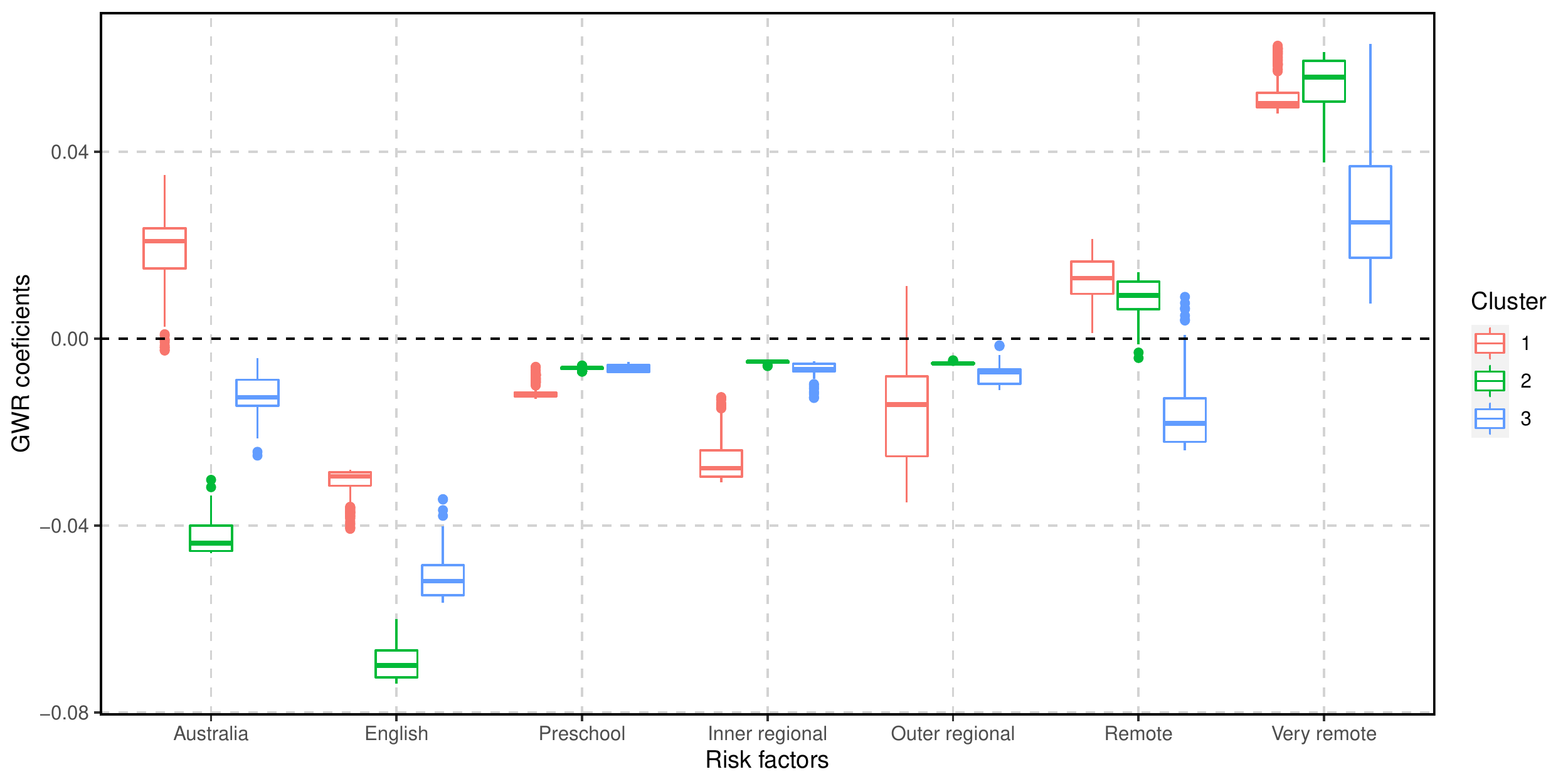}
    \caption{A comparison of the  GWR coefficients for Vuln 1 and risk factors (Australia, English, Preschool,  and remoteness, including inner regional, outer regional, remote, and very remote levels) across the three clusters. }
      \label{fig:my_label6}
      \end{figure}

\begin{figure} [h!]
      \centering
   \includegraphics[width=0.9\textwidth]{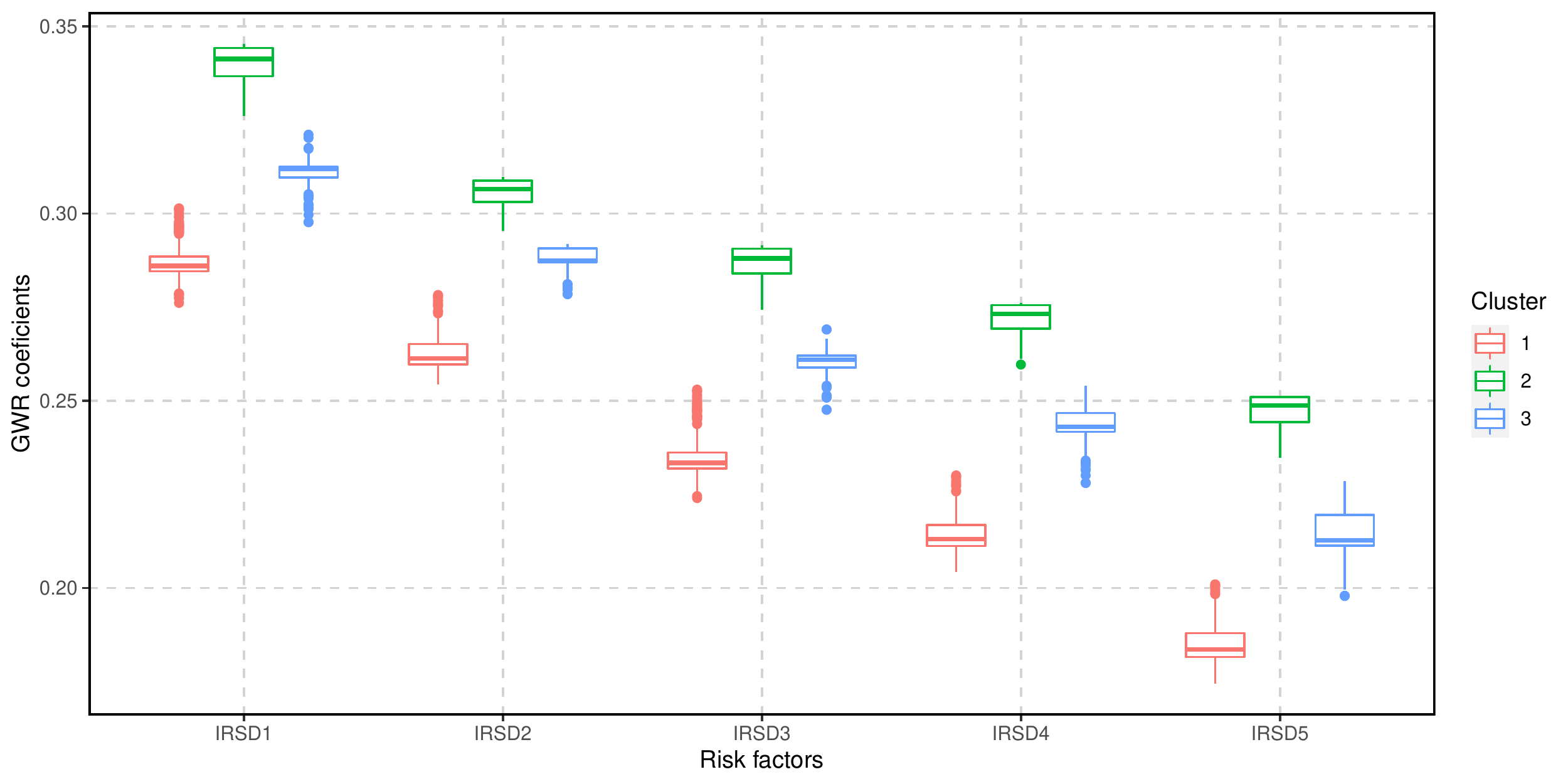}
    \caption{A comparison of the GWR coefficients for Vuln 1 and its associated risk factor, the Index of Relative Socio-economic Disadvantage (IRSD), which has five levels, with level 1 representing the most disadvantaged regions across the three clusters.}
      \label{fig:my_label60}
\end{figure}

The $K$-means cluster analysis of the GWR coefficients for Vuln 1 identified three clusters based on the silhouette coefficients (\nameref{S3}). \figurename{ \ref{fig:my_label5}} shows the distribution of these clusters on the Queensland map.   \noindent  Boxplots of GWR coefficients for areas within each cluster are shown in \figurename{ \ref{fig:my_label6}} and \figurename{ \ref{fig:my_label60}}. In general, cluster 1 was located in the south-east of
Queensland and covered more than half of the state with 342 SA2 areas, including the Greater Brisbane region. Cluster 2 was mainly in
the central west, north-west, northern and far north Queensland, with 104 SA2 areas. Cluster 3 primarily included the southwest and central Queensland
regions with 80 SA2 areas. Cluster 1 had the largest negative GWR coefficients for the proportion of children attending a preschool, indicating that in these areas when the proportion of attendance at preschool increases, the proportions of Vuln 1 decrease. Cluster 2 had the largest negative GWR coefficients for English as the mother language, indicating when the proportion of children with English as the mother language increases, the proportion of Vuln 1 decreases in these regions. Finally, Cluster 3 displayed the greatest GWR coefficients  with inner cities as a baseline, indicating that the proportion of Vuln 1 increases as remoteness increases. \\
In general, all three clusters displayed a negative relationship with preschool attendance. In addition, these clusters show a very high positive association with ``very remote" (compared with inner cities). In addition, these clusters show a relative linear trend that when the disadvantages decrease, Vuln 1 increases.
  \vspace{-1em}
\subsection*{ Summary of the proportion of children attending at preschool for areas in  the first cluster for each type of AEDC domain }

\begin{figure}[h!]
    \centering
    \includegraphics[width=0.8\textwidth]{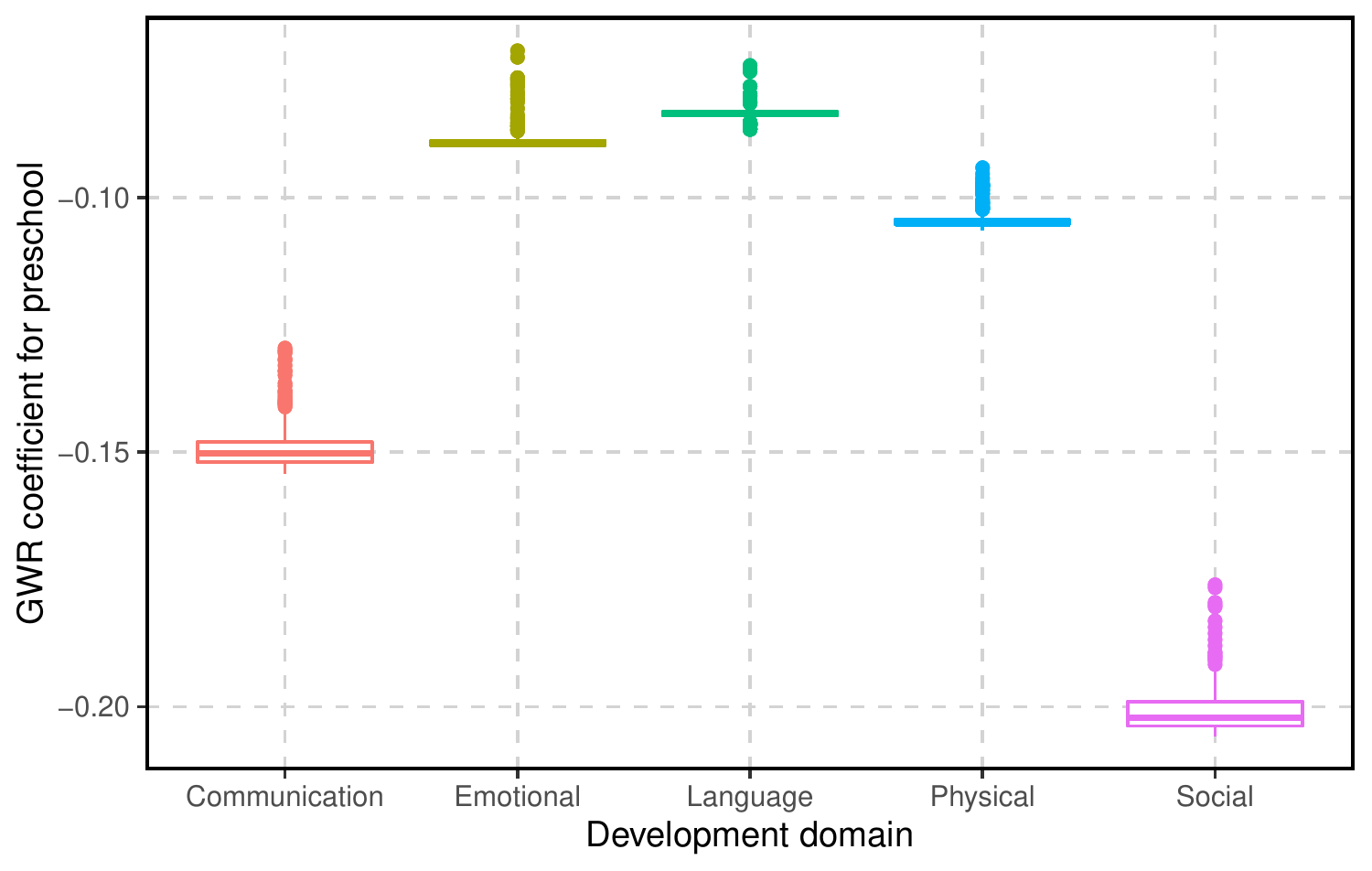}
    \caption{A box plot representation of the coefficients obtained through Geographically Weighted Regression (GWR) for the proportion of attendance at preschool within five domains of the Australian Early Development Census (AEDC) in the first cluster.}
    \label{fig:my_label15}
\end{figure}

Separate GWR and cluster analyses were also performed for each AEDC domain. The results are in the \nameref{support}. We focused on the GWR coefficient corresponding to the proportion of attendance at preschool. In all AEDC domains, the first cluster was found to have the largest negative relationship with the proportion of attendance at preschool. \figurename{ \ref{fig:my_label15}} shows the GWR coefficients for the proportion of attendance at preschool for the five AEDC domains. The figure reveals that preschool has a dominant effect (more negative relationship) for the social competence (Social) and communication skills (Communication) domains, followed by the physical health and well-being (Physical) domain, which shows a focus should be made particularly on improving language and emotional development domains.

\section*{Discussion} \label{disc}
Educational and geographical characteristics have a significant impact on children's development. However, little is known regarding the spatial heterogeneity of the association between these variables and developmental vulnerability in one and more domain(s). This study is one of the first to investigate spatial variation in this association in a large, diverse region but spatially complex, Queensland, Australia,  and to capture spatial clusters of these relationships. While preschool attendance is strongly encouraged in Queensland, the overall participation rate is the lowest among all Australian states and varies geographically.
The analysis found a significant connection between the proportion of
children who attended preschool before they started mainstream school, and the proportion of children measured as developmentally vulnerable in each of the five AEDC domains, as well as in one or more domains (Vuln 1). The study found three distinct clusters inside Queensland.  All three clusters were characterised by a negative mean association between attendance at preschool and Vuln 1, but this relationship varied in consistency within clusters and was affected by different sets of geographic and socio-demographic variables. \\
These results suggest the need for collaboration between health and education partners in an effort to increase preschool access or otherwise improve attendance. Further, health providers may need to consider additional interventions or methodologies for remote areas.\\
These results are consistent with other analyses of the AEDC data. For example, a study found an association between geographic jurisdictions, considering socioeconomic and
demographic characteristics and the probability of
being developmentally vulnerable on one or more 
domain(s) by gender, using nested fixed-effects logistic
regression models \cite{brinkman2012jurisdictional}. Another study investigated the relationship between demographic characteristics and the language
development of children using descriptive statistics, t-tests, one-way analysis of variance (ANOVA) and Tukey multiple
comparison tests \cite{akccay2017examination} . Their study found that the parents' family income and educational background were positively associated with the children's language development. Additionally, another study investigated patterns of universal health and education service use from birth through kindergarten (age four years)  and estimated associations between cumulative risk and service use patterns and between service use patterns and children's developmental vulnerability in the preparatory year (age five years) \cite{taylor2020associations}. The latent class analysis used in this study identified three service use patterns. Membership of low and  high-service user groups was associated with higher cumulative risk and increased odds of developmental vulnerability relative to the regular service user group. These analyses add to recent literature by providing insight into the regions with high vulnerabilities where different policies or investments may be required.\\
There are various limitations to this study. First, the study focuses on 2018 Census data from Queensland. The data for the 2021 census were not available at the time of this study, but it will be interesting to compare results based on this new data set to determine whether these trends are consistent over time and to assess any notable differences. Second, in this study, the data were restricted to the SA2 level, which limited the understanding of spatial patterns and the relationships between vulnerability and covariates at an individual scale. Third, the study only examined educational, socio-demographic and geographical variables. Other variables could be included in future investigations if available at a suitable aggregation level. This study excluded Indigenous status from the GWR model because of local multicollinearity; in particular,  it found a high association between Indigenous status and remoteness level. Since this study focused on geographic variation, it chose to adopt the latter variable. Similarly, there was a high association between socioeconomic factors (IRSD) and Indigenous status (\nameref{S4}). This is supported by other literature \cite{biddle2009ranking,birks2010models}. The complex spatial relationships between Indigenous status, preschool attendance and developmental vulnerability among children  should be considered in separate future work.\\ To identify subgroups among GWR coefficients, various unsupervised clustering methods were used, which showed adequate coverage of common characteristics across algorithms. Unsupervised clustering algorithms are useful for finding subgroups within data without prior knowledge of group labels or classifications \cite{gan2020data}. Different algorithms emphasize different features in their clustering solutions, so analysing data from multiple algorithms is beneficial, especially when there is little prior knowledge of expected subgroups. Comparing results from several methods leads to a more confident identification of well-separated subgroups and reduces sensitivity to method choice. A comparison of three common algorithms ($K$-means, Partition around Medoid, and Hierarchical Clustering) was conducted, and the cluster accuracy can be found in \nameref{S5}. The results show that the three clustering algorithms have a high degree of consistency with accuracy rates above 0.98. This high accuracy suggests strong agreement among the results obtained from each of the clustering algorithms and confirms the robustness of the subgroups identified using different methodologies.
\\

\section*{Conclusion}
Using spatially weighted regression analysis can help to identify the influence of a set of variables on an outcome of interest at each specific geographic location. This study employed spatially weighted regression analysis and $K$-means clustering to investigate at the SA2 level spatial heterogeneity and clustering of the association between the proportion of children attending preschool, taking into account socio-demographic and geographic factors with the proportion of children with developmental vulnerabilities in their first year of full-time school in Queensland, Australia. Three distinct clusters were found with different socio-demographic characteristics. Importantly, the largest cluster revealed a strong negative association between the proportion of attendance at  preschool and the developmentally vulnerable children in their first year of full-time school in Queensland. In these clusters, region-specific interventions may be considered to promote preschool attendance, taking into account socio-demographic issues, in order to minimise development vulnerability among children.




\section*{Supporting information}

\paragraph*{S1 Appendix.}
\label{support}
{\bf Moran's I and local $R^2$.}

\paragraph*{S2 Appendix.}
\label{S2}
{\bf Clusters inside Greater Brisbane and Summary of GWR coefficients.}

\paragraph*{S3 Appendix.}
\label{S3}
{\bf Additional analysis.} 
A: Silhouette score. B: GWR coefficients for each type of AEDC domain
\paragraph*{S4 Appendix.}
\label{S4}
{\bf Relation between Indigenous and other socio-demographic variables.}

\paragraph*{S5 Appendix.}
\label{S5}
{\bf Cluster Accuracy.} 

\section*{Acknowledgments}
We would like to express our gratitude to the team at Children Health Queensland and the Center for Data Science for their invaluable assistance and support in this project.

\nolinenumbers

%
%
%

\end{document}